\documentclass[a4paper,conference]{IEEEtran}
\usepackage{cite}
\ifCLASSINFOpdf
  \usepackage[pdftex]{graphicx}
  \graphicspath{{../pdf/}{../jpeg/}}
  \DeclareGraphicsExtensions{.pdf,.jpeg,.png}
\else
\fi
\usepackage{amsmath}
\renewcommand\IEEEkeywordsname{Keywords}
\usepackage{enumitem}
\usepackage{dblfloatfix}
\usepackage{amsfonts}

\usepackage{xcolor}
\newcommand{\revised}[1]{{#1}}

\IEEEoverridecommandlockouts
\begin{document}
\title{Semantic Parsing of Interpage Relations}

%\author{\IEEEauthorblockN{Mehmet Arif Demirta\c{s}}
%\IEEEauthorblockA{Department of Computer Engineering\\
%Istanbul Technical University\\
%Istanbul, Turkey\\
%Email: demirtasm18@itu.edu.tr}
%\and
%\IEEEauthorblockN{Berke Oral}
%\IEEEauthorblockA{Yap{\i} Kredi Teknoloji\\
%Istanbul, Turkey\\
%Email: berke.oral@ykteknoloji.com.tr}
%\and
%\IEEEauthorblockN{Mehmet Yasin Akpınar}
%\IEEEauthorblockA{Yap{\i} Kredi Teknoloji\\
%Istanbul, Turkey\\
%Email: mehmetyasin.akpinar@ykteknoloji.com.tr}
%\and
%\IEEEauthorblockN{Onur Deniz}
%\IEEEauthorblockA{Yap{\i} Kredi Teknoloji\\
%Istanbul, Turkey\\
%Email: onur.deniz@ykteknoloji.com.tr}}


%\author{\IEEEauthorblockN{Mehmet Arif Demirta\c{s}\IEEEauthorrefmark{1}\IEEEauthorrefmark{2},
%Berke Oral\IEEEauthorrefmark{2},
%Mehmet Yasin Akpınar\IEEEauthorrefmark{2} and
%Onur Deniz\IEEEauthorrefmark{2}}
%\IEEEauthorblockA{\IEEEauthorrefmark{1}Department of Computer Engineering, Istanbul Technical University, Istanbul, Turkey}
%\IEEEauthorblockA{\IEEEauthorrefmark{2}Yap{\i} Kredi Teknoloji, Istanbul, Turkey}
%\IEEEauthorblockA{Email: demirtas18@itu.edu.tr, \{berke.oral, mehmetyasin.akpinar, onur.deniz\}@ykteknoloji.com.tr}}

\DeclareRobustCommand*{\IEEEauthorrefmark}[1]{%
  \raisebox{0pt}[0pt][0pt]{\textsuperscript{\footnotesize\ensuremath{#1}}}}
\author{\IEEEauthorblockN{Mehmet Arif Demirta\c{s}\IEEEauthorrefmark{1,}\IEEEauthorrefmark{2},
Berke Oral\IEEEauthorrefmark{2}, %\thanks{*Work done while the authors were at Yapi Kredi Teknoloji.}
Mehmet Yasin Akpınar\IEEEauthorrefmark{2} and
Onur Deniz\IEEEauthorrefmark{2}}
\IEEEauthorblockA{\IEEEauthorrefmark{1}Department of Computer Engineering, Istanbul Technical University, Istanbul, Turkey}
\IEEEauthorblockA{\IEEEauthorrefmark{2}Yap{\i} Kredi Teknoloji, Istanbul, Turkey}
\IEEEauthorblockA{Email: \{demirtas18, oralbe\}@itu.edu.tr, \{mehmetyasin.akpinar, onur.deniz\}@ykteknoloji.com.tr}}


\maketitle

\begin{abstract}
Page-level analysis of documents has been a topic of interest in digitization efforts and multimodal approaches have been applied to both classification and page stream segmentation. In this work, we focus on capturing finer semantic relations between pages of a multi-page document. To this end, we formalize the task as \textit{semantic parsing of interpage relations} and we propose an end-to-end approach for interpage dependency extraction, inspired by the dependency parsing literature. We further design a multi-task training approach to jointly optimize for page embeddings to be used in segmentation, classification, and parsing of the page dependencies using textual and visual features extracted from the pages. Moreover, we also combine the features from two modalities to obtain multimodal page embeddings. To the best of our knowledge, this is the first study to extract rich semantic interpage relations from multi-page documents. Our experimental results show that the proposed method increased LAS by 41 percentage points for semantic parsing, increased accuracy by 33 percentage points for page stream segmentation, and 45 percentage points for page classification over a naive baseline.

%we formalize the task as the semantic 

%Our experimental results show that our method achieves comparable results on page stream segmentation with the existing work while also extracting the page relations to be used in the downstream tasks.

%Segmentation of document streams obtained from digitization of multi-page documents have been studied in detail. However, previous works ignore possible relations between the segmented documents obtained from the same stream, causing a loss of potential information. In this work, we define \textit{page dependencies} to capture finer semantic relations between segmented documents. Inspired from the dependency parsing literature, we propose the use of a transition-based parser for page dependency extraction and we further design a multi-task training approach to jointly optimize for document embeddings to be used in segmentation, classification and parsing of the page dependencies. Our experimental results show that our method achieves comparable results on page stream segmentation with the existing work with significantly less data while also extracting the page relations to be used in the downstream tasks. 
\end{abstract}

\vspace{1em}
\begin{IEEEkeywords}
semantic parsing, page stream segmentation, dependency parsing, multimodal page representation, document understanding
\end{IEEEkeywords}

% \fi
\IEEEpeerreviewmaketitle

\section{Introduction}
% This demo file is intended to serve as a ``starter file''
% for IEEE conference papers produced under \LaTeX\ using
% IEEEtran.cls version 1.8b and later.
% I wish you the best of success.



Digital processing of printed documents is a major part of the workflow of multiple industries, including banks, customs, and judicial institutions. Information extraction from digitized documents on a page-by-page basis is thoroughly examined in the natural language processing (NLP) literature~\cite{nadeau_survey_2007, nguyen_relation_2015}. Multiple studies focus on the task of document image classification~\cite{dong_document_2008, rusinol_multimodal_2014, awal_complex_2017, jain_multimodal_2019}. In addition to analysis of individual pages, \textit{page stream segmentation} (PSS) task considers a continuous stream of pages obtained from batch-scanning of multiple documents, where the objective is to break the stream into subsets of pages by detecting the pages that correspond to the first page of a separate document in the stream~\cite{gordo_document_2013}. The segmentation task is traditionally done via seperator sheets inserted into streams by human operators, but this process is both error-prone and costly with an error rate up to 8\% and takes up to 50\% of the total document preparation cost~\cite{schmidtler_automatic_2007}. Later works on PSS incorporate models with multiple components and deep learning methods in order to achieve higher segmentation accuracy~\cite{daher_multipage_2014, gallo_deep_2016, wiedemann_multi-modal_2021}.

Despite the developments on segmentation and classification, semantic relations \textit{between} pages have not been taken into consideration. Current works assume the segmented documents to be completely independent from each other, where in many real-world applications, a batch-scanned stream of pages is made of subdocuments that are related with each other, such as attachments or multiple copies of a document in the same stream. Extraction of these semantic relations in addition to segmentation info can speed up document processing workflows and help improve the accuracy of downstream NLP tasks by providing extra information.

For this study, we focused on the domain of international trade documents, which consist of many subdocuments related to each other. These documents are supplied to the entities such as customs and banks in order to legalize the international money transfers which are mainly conducted by the SWIFT system. The subdocuments are listed in the SWIFT message for the relevant transaction and they need to be verified. The verification step includes checking the consistency of the information throughout the subdocuments and whether the listed subdocuments are submitted in the correct counts (some subdocuments, e.g. invoice and bill of lading, are expected to be submitted more than one copy and contain exactly the same information). Therefore, it is crucial for these relations to be extracted in order to fully automatize the process.

% The verification step consists of checking if the listed subdocuments are sent in the correct counts. 

We define \textit{page dependencies} to capture these relations inspired by dependency parsing literature, such that a dependent page $p_i$ is related to a head page $p_j$ with a dependency label $r$ for $i, j \in [1, N]$ for a document of $N$ pages. Using this notation, a document can be represented with a set of $N$ page dependencies. We capture the pages that are attachments, copies, back pages or next pages of another page with this representation. To parse these semantic dependencies, we propose to use a transition-based dependency parser with a neural classifier for choosing transitions. By tokenizing a document into pages and creating feature embeddings for each token instead of tokenizing sentences into words, we show that a traditional parser can easily be adapted to be used on a document level. We propose multimodal feature embeddings that capture both visual and textual information from the pages. 
It should be noted that the proposed method is not bound to our set of 4 interpage relation labels, and it can be trained to recognize other interpage dependency structures.

We further propose a multi-task training procedure for optimizing feature embeddings by jointly training \revised{an embedding module and three task specific modules}. The first \revised{task-specific} module is a PSS classifier that classifies each page as the start of a new document, continuation of the previous page, or an empty page for detecting defects/human errors in the scanning process. The second module is the aforementioned dependency parser that assigns each dependency one of four dependency labels, and the third module is a fine scale classifier that assigns each page one of 33 semantic classes. We test our models on our in-house dataset of trade documents, collected from 146 batch-scanned streams consisting of 5345 pages in total.

In summary, the main contribution of our work is the application of dependency parsing algorithms for extraction of relations between pages. To the best of our knowledge, there have not been any work on extracting semantic relations between pages on multi-page documents. In addition to the formulation of semantic parsing for relations between pages, we provide a multimodal multi-task setup for document processing workflows for segmentation, classification and relation extraction from the pages. Our parsing methods achieve promising performance with unlabeled and labeled attachment scores of up to $77.6\%$ and $71.2\%$, improving our baseline results by 29 and 41 percentage points respectively. Proposed system also achieve segmentation performance close to the human error rate of $8\%$ with $92.2\%$ F1 score, and improves classification accuracy by 45 percentage points over our baseline and 5 percentage points over unimodal embeddings.

The remainder of this paper is organized as follows. We cover the related works on page stream segmentation and multimodal document processing in Section II, detail the formulation of page dependencies and our model design in Section III, explain our experimental setup and discuss our results in Section IV and conclude our findings in Section V.

\section{Related Works}
\textbf{Page Stream Segmentation.} 
Initial works on segmentation of document streams focused on probabilistic approaches. \cite{collins2002clustering} presented the task as a clustering problem and performed bottom-up clustering based on a page similarity measure derived from handcrafted features including page numbers, header/footer similarity and overall layout/text similarity between pages. \cite{schmidtler_automatic_2007} described the task as a sequence mapping problem, and used support vector machine classifiers on features extracted from the text using bag-of-words model to obtain classification probabilities, which are then used in a finite state transducer-based approach for boundary detection. \cite{meilender_segmentation_2009} proposed a multi-gram based model inspired from variable horizon models used in speech recogniton.

\cite{gordo_document_2013} uses multi-page document representations to learn a probabilistic model of document validity and leverages textual descriptors to propose the segmentations that obtain the maximal validity scores. \cite{daher_multipage_2014} employs a two-step model where document candidates are formed through a segmentation module that classifies the pages as either continuation of the previous page or a new page, and a verification module predicts uncertainity scores for the formed segmentations using bag-of-words features. \cite{hamdi_machine_2017} compares Doc2Vec representations of documents to rule-based segmentation approaches and shows that machine learning models can outperform traditional methods in multi-page datasets. \cite{gallo_deep_2016} is the first work to incorporate visual features extracted by convolutional neural networks and uses an additional deep neural network that takes the probability distributions obtained via CNN as input in a sliding window manner to use the information for neighbouring pages in classification and segmentation.

More recently, \cite{neche_use_2020} showed recurrent neural networks combined with attention layers outperforms other text-based approaches at segmentation including handcrafted features and Doc2Vec representations. \cite{wiedemann_multi-modal_2021} introduces multimodality by early and late fusion techniques. Their text embeddings are extracted through fastText embeddings and Bi-GRU recurrent layers, whereas their visual embeddings are obtained using a pretrained VGG16 network with only the last block being trainable. \cite{braz_leveraging_2021} frames the problem as a classification of 4 classes that can be reduced to the traditional 2 class problem in the literature. They also investigate the use of more computationally efficient models compared to previous work. While page stream segmentation literature examined the use of a variety architectures and modalities, this has been the only work that detaches from the traditional binary classification framework, and more detailed semantic approaches have not been a topic of consideration to the best of our knowledge.

\textbf{Multimodal Page Representations.}
Combination of textual and visual modalities has proved to be beneficial in a multitude of NLP tasks. \cite{liparas_news_2014} achieves consistent performance increases on classification of web articles by using N-gram textual features and visual descriptors. \cite{rusinol_multimodal_2014} presents early and late fusion techniques for combining visual features obtained from pixel intensity distributions and textual features obtained with bag-of-words model for page classification. \cite{cristani_multimodal_2016} uses multimodal vector representations to analyze the similarities of documents in vector spaces. \cite{audebert_multimodal_2019} uses a pretrained convolutional neural network (CNN) for visual feature extraction and word embeddings for textual feature extraction, which are fused through concatenation. Their classification results show that multimodal approach outperforms textual modality by a margin and slightly improves visual modality. \cite{engin2019multimodal} proposes both early and late fusion techniques for multimodal classification of banking documents. Their early fusion network uses LSTM layers to obtain text embeddings which are then concatenated with pretrained CNN output for visual embeddings, whereas late fusion network concatenates the probability distributions obtained from two unimodal networks as the input of a classification module. Both fusion techniques improved over unimodal approaches on classification. Moreover, they showed that fine-tuning of the pretrained visual network in early fusion increased accuracy. \cite{jain_multimodal_2019} builds upon earlier works with four early fusion techniques. In addition to summation, concatenation and outer products, they adapt multimodal gated units from \cite{arevalo_gated_2017} to document classification domain where the contribution of each modality into the multimodal representation is controlled by a learnable gating mechanism based on the input modalities. 

\section{Methodology}

\begin{figure}
    \centering
    \scalebox{1}{
    \includegraphics[width=0.48\textwidth, trim={1.8cm 2.73cm 3cm 1.5cm}, clip]{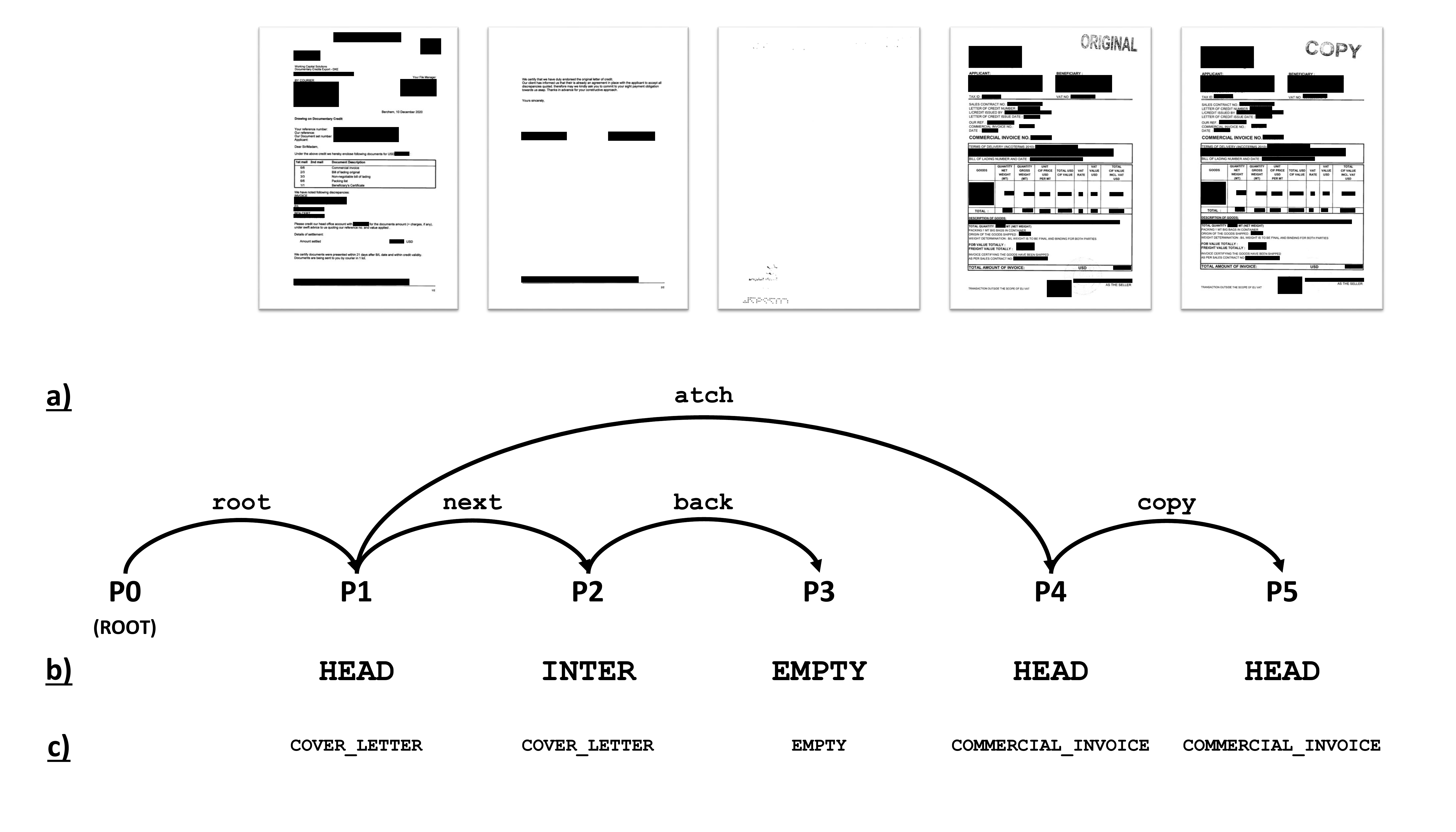}
    }
    \caption{An example document with ground truth labels. a) shows the dependency tree where each page is mapped to a token with a dummy root token, and each relation is labeled with an arc label. b) shows the segmentation tags assigned for the page stream segmentation task. c) shows the semantic tags assigned for the page classification task. For this example, the multi-page document contains three subdocuments; a cover letter with three pages, where the last page is empty and mistakenly scanned, a commercial invoice attached to the cover letter, and a copy of the same commercial invoice. Sensitive information is blacked out for privacy purposes.}
    \label{Fig:Document}
\end{figure}

In this section we explain our formulation of page dependencies, creation of multimodal feature embeddings from pages, and our model architecture for multi-task learning.

\subsection{Semantic Parsing of Page Dependencies}
To parse the interpage relations of a document, we first formally express these relations as page dependencies. We build our page dependency formulation on the existing dependency parsing literature~\cite{kubler_dependency_2009}. We denote a document $D$ as a sequence of tokens $ p_0p_1p_2\ldots p_N$ where $p_0$ is an artificial token for the root of the dependency tree that do not depend on any other token, and rest of the tokens correspond to the pages of the document. We adapt \textit{dependency relation types} (also known as \textit{arc labels}) where the set of types is denoted as $R = \{r_1,\ldots,r_m\}$ and $r_i$ is the semantic meaning from between two tokens. Using these two definitions, we can formally define a page dependency as a 3-tuple $(p_i, r, p_j)$ such that there is a dependency relation from head page $p_i$ to dependent page $p_j$ with the dependency label $r \in R$. From the page dependency definition, we can represent a document $D$ with the corresponding dependency tree $G = (V, A)$ with set of vertices $V$, set of edges $A$ and root $p_0$:
\begin{enumerate}[itemsep=-4mm]
    \item $V \subseteq {p_0,\ldots,p_N}$\\
    \item $A \subseteq V \times R \times V$\\
    \item $(p_i, r, p_j) \in A\; \forall p_j \in V\setminus\{p_0\}$\\
    \item $(p_i, r, p_0) \not\in A$\\
    \item if $(p_i, r, p_j) \in A$ then $(p_i, r', p_j) \not\in A$ for all $r' \neq r$\\
    \item if $(p_i, r, p_t) \in A$ then $(p_j, r, p_t) \not\in A$ for all $i \neq j$
\end{enumerate}

To better demonstrate the connection from the document to the dependency tree, Figure~\ref{Fig:Document} shows an example document, its page dependencies and the visualization of the dependency tree. The given example also shows the arc labels for our dataset:
\begin{itemize}
    \item \texttt{atch} connects a head page to its attachment
    \item \texttt{copy} connects a head page to its copy
    \item \texttt{back} connects a head page to its back page
    \item \texttt{next} connects a head page to its continuation
\end{itemize}

Documents connected by \texttt{back} and \texttt{next} are part of the same document segment, where \texttt{atch} and \texttt{copy} connect pages representing the first pages of different document segments. 

Page dependency notation enables the use of any data-driven dependency parsing model to extract the dependency tree for a document, by forming a similarity between semantic parsing of interpage relations and dependency parsing of sentences. In this setup, both tasks can be accomplished with the same model architecture given that pages of the document can be represented with feature vectors similar to tokens of a sentence. 

% In this setup, we extract feature vectors for pages similar to the feature vector extraction for words in dependency parsing literature. By forming a similarity between semantic parsing of interpage relations and dependency parsing of sentences we can use  any data-driven dependency parsing model to extract the dependency tree for a document. Both tasks can be accomplished with the same model architecture given that pages of the document can be represented with feature vectors similar to tokens of a sentence.

%\begin{figure}
 %   \centering
  %  \includegraphics[width=0.44\textwidth, trim={0 3cm 7cm 2cm}, clip]{figures/fig4.pdf}
   % \caption{An example of similar visual layout between two copy pages. Both pages share the same header except for the tag ORIGINAL/COPY (1), both include a complex table (2) and the stamp and signature of the authorized figure in slightly different positions (3).}
%    \label{fig:VisualLayout}
%\end{figure}

\subsection{Token to Feature Embedding Conversion}
As explained, we represent each document as a sequence of tokens with feature embeddings, where each token corresponds to a page of the document. We hypothesize that both visual and textual features can help identify the semantic relations that are captured through page dependencies. For instance, textual content of a page labeled with arc label \texttt{next} usually contain references to the previous page or page numbering, whereas pages labeled with arc label \texttt{copy} share the same visual layout but their text may differ due to the noise in OCR processes. A detailed example showing contributions of both modalities is shown in Appendix A.

Therefore, we first design visual and textual feature extraction methods and we also combine these unimodal feature embeddings with a variety of fusion techniques to obtain multimodal embeddings that captures the information from both channels in an adequate way.

\textbf{Visual Embeddings.} To extract visual features from a page, we use a convolutional neural network (CNN) on the downscaled page images followed by a dropout layer and two linear layers, with a nonlinearity at the end of first linear layer. Specifically, we use MobileNetV2 architecture~\cite{sandler_mobilenetv2_2019} as the CNN due to its success on image classification tasks and efficiency. 

\textbf{Textual Embeddings.} \revised{To obtain textual embeddings, we first use a Java SDK of an OCR product to obtain text from the documents which are in image formats initially. \footnote{The OCR product used for the process is ABBYY FineReader Engine. This product is chosen because it was already in use for the document processing pipeline prior to this study. Detailed parameters are given in Appendix B.} After obtaining the OCR text of the documents, we use the word vectors of the first 512 tokens from the full text of each page as the input to a BiLSTM layer, followed by a dropout and an attention layer, and a final linear layer.} We use a set of custom word vectors trained on banking documents to map tokens to word vectors.

\textbf{Multimodal Embeddings.} To obtain the multimodal embeddings for our tasks, we apply early fusion by combining the outputs of textual and visual embeddings before feeding the vectors into task-specific subnetworks. In addition to vector based fusion methods, we used weighted addition approaches from \cite{oral2022fusion}.
For combining textual embedding $\mathbf{w}^t \in \mathbb{R}^{N}$ and visual embedding $\mathbf{w}^v  \in \mathbb{R}^{N}$ into multimodal embedding $\mathbf{w}^m$, we utilize five fusion methods:

\begin{itemize}
    \item Concatenation: \begin{gather*}\mathbf{w}^m = \mathbf{w}^t \mathbin\Vert \mathbf{w}^v \textrm{ where } \mathbf{w}^m \in \mathbb{R}^{2N} \end{gather*}
    \item Element-wise sum: \begin{gather*}\mathbf{w}^m_i = \mathbf{w}^t_i + \mathbf{w}^v_i \textrm{ where } \mathbf{w}^m \in \mathbb{R}^{N}\end{gather*}
    \item Weighted element-wise sum, with learnable scalar weight 
    \begin{multline*}\mathbf{w}^m_i = \lambda\mathbf{w}^t_i + (1-\lambda)\mathbf{w}^v_i\\ \textrm{ where } \mathbf{w}^m \in \mathbb{R}^{N} \land \lambda \in \mathbb{R}\end{multline*}
    \item Weighted element-wise sum, with learnable weight vector
    \begin{multline*}\mathbf{w}^m_i = \Lambda_i\mathbf{w}^t_i + (1-\Lambda_i)\mathbf{w}^v_i\\ \textrm{ where } \mathbf{w}^m \in \mathbb{R}^{N} \land \Lambda \in \mathbb{R}^N\end{multline*}
    \item Weighted element-wise sum, with weight as a learnable function of page contents
     \begin{multline*}\mathbf{w}^m_i = \lambda\mathbf{w}^t_i + (1-\lambda)\mathbf{w}^v_i\\ \textrm{ where } \mathbf{w}^m \in \mathbb{R}^{N} \land \lambda = f(\mathbf{w}^v) \land f: \mathbb{R}\to\mathbb{R}\end{multline*}
\end{itemize}

In order to capture the information encoded in neighbouring pages, we form a window for each embedding by concatenating vectors of neighbouring tokens with itself. We then apply maxout activation followed by residual connections as proposed in ~\cite{honnibal_spacy_2017} in order to encode context to the obtained feature vectors.
%We also propose a multi-task model architecture for the training of these embeddings. Next section will detail our model approach.

\subsection{Multi-task Model Architecture}

\begin{figure}
    \centering
    \includegraphics[width=0.48\textwidth, trim={0 6cm 7cm 3cm}, clip]{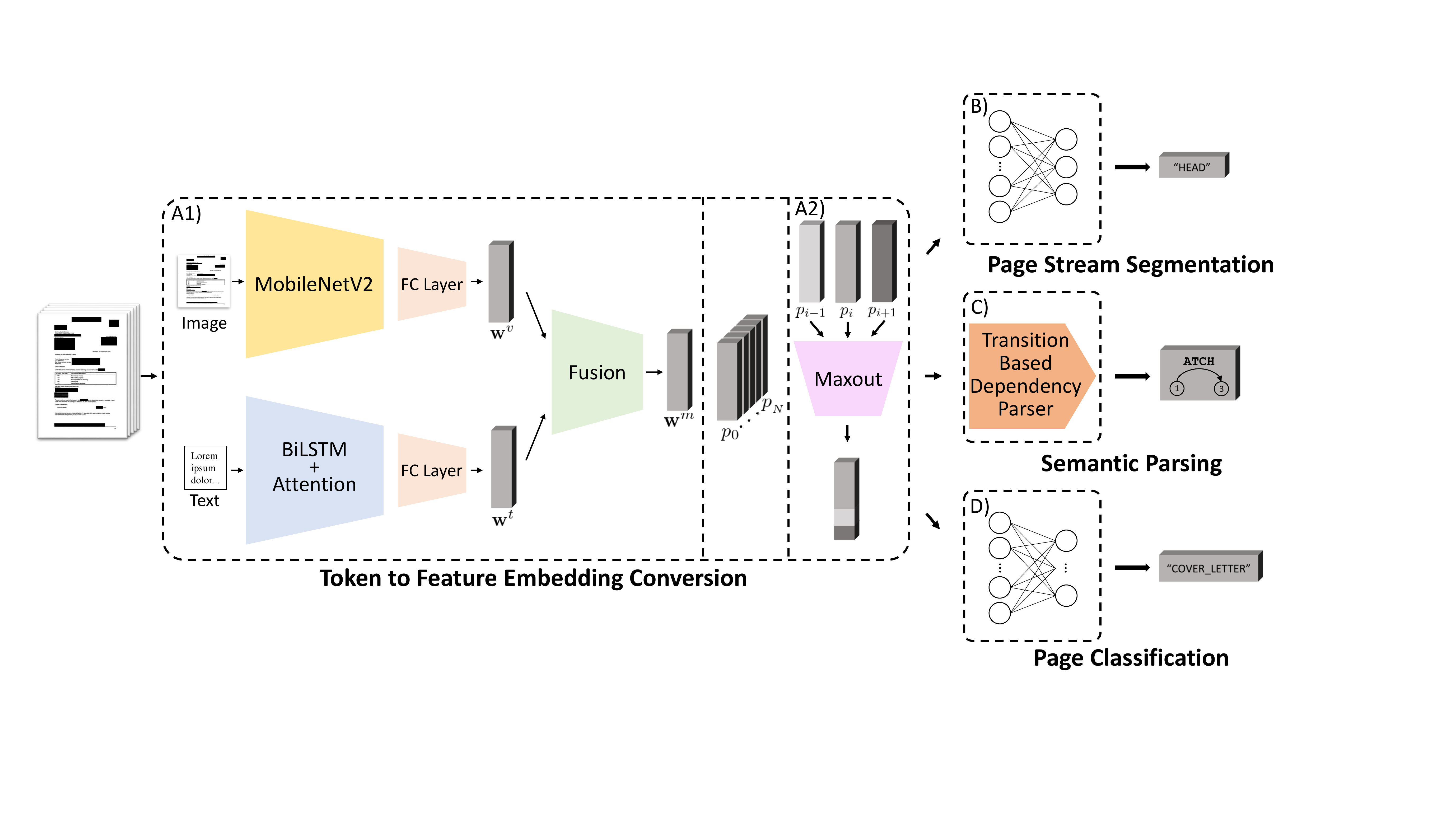}
    \caption{Proposed model architecture. First, text and image content of the pages are used to obtain and fuse unimodal embeddings (A1) for each token, and these embeddings go through a maxout activation with the neighbouring vectors to encode context information (A2). Obtained embeddings are used in page stream segmentation (B) as the input of a softmax layer to predict segmentation tags, in semantic parsing (C) as the input of a neural transition-based dependency parser to predict relations between pages, and in page classification (D) as the input of a softmax layer to predict page classes.}
    \label{Fig:Model_Architecture}
\end{figure}

Our proposed model consists of four main modules. The first module handles the token to feature vector conversion detailed in previous subsection, whereas the remaining three modules are specific to each task and work in parallel for segmentation, dependency parsing and classification respectively. An overview of our architecture can be seen in Figure~\ref{Fig:Model_Architecture}. Our design is inspired by the pipeline structure of SpaCy framework~\cite{honnibal_spacy_2017}, \revised{where components are combined in a processing pipeline similar to our modules for processing documents}.

\textbf{Page Stream Segmentation.} This module breaks the multi-page document into subdocuments by predicting a segmentation tag for each page. We use three segmentation tags: $\texttt{HEAD}$ for the start of a new document, $\texttt{INTER}$ for the continuation of the previous page and $\texttt{EMPTY}$ for the pages that do not have any meaningful content but considered as the continuation of the previous page. These tags are also shown in Figure~\ref{Fig:Document}b.
\revised{The prediction of these tags are done via a linear layer with softmax activation. This layer is trained with categorical crossentropy loss to map input embeddings obtained from the token to feature embedding conversion module explained above into segmentation tags.}
%To predict these tags, a linear layer with softmax activation is trained with categorical crossentropy loss to map input embeddings to segmentation tags. 

\textbf{Dependency Parsing.} This module parses the page dependencies into a tree representation. For each token, an arc label and a dependency head index is predicted using a transition-based dependency parsing algorithm that uses neural networks for creating state representations from input tokens and predicting scores for possible transitions.
\revised{Our model uses the architecture from SpaCy~\cite{honnibal2015improved}, consists of a lower network of hidden size $w$ that computes the intermediate matrices of shape $(d, w)$ for each token where $d$ is the embedding dimension. Then, a maxout nonlinearity is applied, followed by an upper network, which is a linear layer to predict scores for each state generated by the parser. Note that the parsing module is independent from the page embedding creation and could be replaced with an alternative parsing approach, such as graph-based parsers by only modifying the source of node embeddings~\cite{ji2019graph}.}

\textbf{Page Classification.} This module classifies each page of the document into one of 33 semantic classes. This classes are based on the contents of the pages, as shown in Figure~\ref{Fig:Document}c. 
\revised{This module also uses a linear layer with softmax activation to map input embeddings learned for each page to document classes. The architecture is very similar to page stream segmentation with the only change being the number of output tags/classes.}
%Similar to page stream segmentation, this component uses a linear layer with softmax activation to classify input embeddings and uses categorical crossentropy loss for training.

\section{Results \& Discussion}

\subsection{Dataset}
\begin{figure}[ht]
    \centering
    \scalebox{0.8}{
    \includegraphics[width=0.5\textwidth]{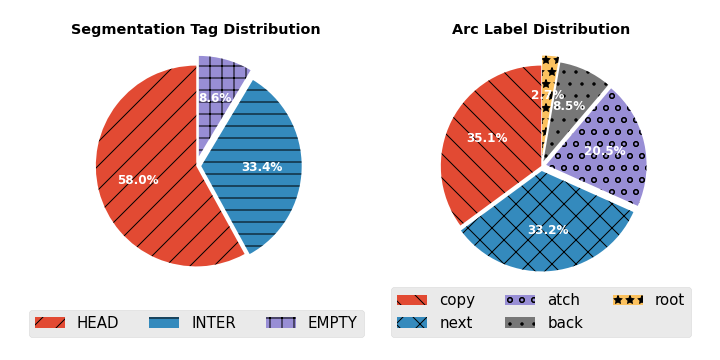}
    }
    \caption{Frequency of segmentation tags (left) and arc labels (right)}
    \label{Figure:Dist_Data}
\end{figure}

We created and annotated a new dataset for the evaluation of our methods.\footnote{Keeping the subtasks in mind, a research for the public datasets is conducted during the study. The most important feature for the dataset for this work is having multipage documents that are consisting of subdocuments. The closest dataset that we could find is Tobacco800 database \cite{lewis2006building}, since it contains multipage document images splitted into single page files. However, there are no subdocuments or relations between documents present in the dataset, which makes it unfit for the study.} Our dataset includes 146 multi-page documents collected from international trade documents with 5345 pages in total, with each document containing $36.6$ pages on average with a standard deviation of $18.2$. Shortest and longest documents contain 4 and 82 pages, respectively. Each page is labeled with a segmentation tag, an arc label, a dependency head index (pointing to the parent of the node in the dependency tree), and a semantic class label. Figure~\ref{Figure:Dist_Data} shows the distribution of segmentation tags and arc labels. Each document contains subdocuments that we try to segment and relate to each other through page dependencies. In total, there are 3101 subdocuments in the dataset with mean page count $1.724 \pm 1.141$, with longest subdocument containing 11 pages. Table~\ref{Table:Dist_Page} shows the distribution of page counts of subdocuments. We are unable to share the dataset as it contains financially confidential information.

\vspace{-1.5em}

% \begin{table}[h]
% \renewcommand{\arraystretch}{1.3}
% \begin{tabular}{c|c}
% \hline
% \textbf{Segmentation Tag} & \textbf{Percentage} \\ \hline
% HEAD & 0.580 \\ 
% INTER & 0.334 \\ 
% EMPTY & 0.086 \\ \hline
% \end{tabular}
% \caption{Distribution of segmentation tags in the dataset}
% \label{Table:Dist_Segment}
% \end{table}
% \begin{table}[h]
% \renewcommand{\arraystretch}{1.3}
% \begin{tabular}{c|c}
% \hline
% \textbf{Arc Label} & \textbf{Percentage} \\ \hline
% copy & 0.351 \\ 
% next & 0.332 \\
% atch & 0.205 \\ 
% back & 0.085 \\ 
% root & 0.027 \\ \hline
% \end{tabular}
% \caption{Distribution of arc labels in the dataset}
% \label{Table:Dist_Arc}
% \end{table}

\begin{table}[h]
\centering

\renewcommand{\arraystretch}{1.3}
\caption{Distribution of subdocuments for Page Stream Segmentation}
\scalebox{0.80}{
\begin{tabular}{c|c}
\hline
\textbf{Subdocument length} & \textbf{Count} \\ \hline
1 & 1780 \\ 
2 & 845 \\ 
3 & 193 \\
4 & 208 \\ 
5 & 26 \\ 
6 & 33 \\
%7 & 4 \\ 
%8 & 7 \\ 
%9 & 0 \\
%10 or more & 5 \\\hline
7 or more & 16 \\\hline
%Single-page subdocuments & 1780 ($57\%$)\\
%Multi-page subdocuments & 1321 ($43\%$)\\\hline
%Pages in new document & 3101 ($58\%$)\\
%Pages in same document & 2244 ($42\%$)\\\hline
Total subdocuments & 3101\\
Total pages & 5345\\\hline
\end{tabular}
}
\label{Table:Dist_Page}
\end{table}

\subsection{Implementation Details}
Explained methods are implemented using SpaCy framework~\cite{honnibal_spacy_2017}. While SpaCy enables users to apply dependency parsing, tagging or classification modules on sentences, it does not provide a way to use these modules on documents. Therefore, we modified the feature extraction steps that normally map the linguistic features of sentences to hashes to be used in the later layers to instead load the visual and textual content of documents and extract embeddings from them using the methods explained in the previous section. By mapping document pages to tokens, it is possible to use higher level processing modules such as dependency parsing directly on the documents.

For the visual content, 3-channel RGB page images are resized to 224 by 224 and normalized to the range $[0, 1]$. While the provided document files are in RGB color space, in reality documents do not contain visual content in color. We chose not to convert the files to grayscale as the pretrained networks we used are also trained on RGB images. We use 96-dimensional vectors for page embeddings. In visual embeddings, input dimension of the last linear layer is 512.

\revised{For the textual content, first the OCR texts are obtained, then the textual embeddings are created.
%\begin{itemize}
%    \item CorrectInvertedImage: true
%    \item FastMode: true
%    \item LowResolutionMode: true
%    \item DetectTables: true
%    \item EnableTextExtractionMode: true
%    \item ProhibitModelAnalysis: true
%    \item DetectPictures: false
%    \item DetectTextOnPictures: true
%    \item TextTypes: TT\_Normal $\vert$ TT\_Typewriter $\vert$ TT\_OCR\_B
%    \item GeometryCorrectionMode: GCM\_Auto
%    \item CorrectSkew: True
%    \item CorrectSkewMode: CSM\_CorrectSkewByHorizontalLines $\vert$ CSM\_CorrectSkewByHorizontalText
%    \item The rest of the parameters are left in default values.
%\end{itemize}
These embeddings consist of 100-dimensional word vectors.} The word vectors are trained with fastText \cite{bojanowski2016enriching} algorithm using corpora of foreign trade documents. For out of vocabulary words, we assign zero vector instead of using character n-grams to create representation. \revised{Linear layers for page stream segmentation and page classification map these 96-dimensional vectors to one of the 4 or 33 classes respectively through a softmax layer. For the dependency parsing module, the lower network uses a hidden size of 64 and maxout nonlinearity uses 2 pieces in the state prediction layer. The parameters were assigned manually based on our early empirical results.}

We trained each model for 200 epochs with batch size set to 4. We used Adam optimizer decoupled weight decay as proposed in \cite{loshchilov_decoupled_2019} with learning rate set to 0.001 and weight decay set to $10^{-6}$. 

\subsection{Experimental Results}
Table~\ref{Table:Result_Embedding} shows the results for each of the three tasks using different embedding modalities. We evaluate page stream segmentation performance with the accuracy of segmentation tags and F1 score of \texttt{HEAD} tag, corresponding to the continuity/rupture classification in previous works. For classification, we report accuracy and macro F1 score. For semantic parsing, we report unlabeled (UAS) and labeled (LAS) attachment scores, which denotes the ratio of tokens assigned to the correct head token without considering the label (UAS) or with the correct label (LAS). 3-fold cross validation is performed and metrics are calculated over the combined predictions of each fold for each embedding creation method.

To compare our results, \revised{we defined a naive baseline method using the indices of the pages.\footnote{We employ the default hash embedding model from SpaCy on page indices. Each index is transformed into a hash, and the hashes are used in an embedding table to obtain an embedding vector for each page. The width of these embedding vectors is set to be the same as the embeddings from our proposed methods.} Naive embeddings use only the location of the page inside the document. As semantic parsing is a novel task, we use this baseline to compare our proposed methods against a model which does not access textual and visual features from the pages.}
%The only information contained in the embeddings is the location of the page inside the document. As no tasks similar to our semantic parsing definition has been covered in the existing literature, we essentially create this baseline to compare our proposed embedding methods against a model which does not have access to visual and textual features of the pages and only have access to the page number information. The rest of the architecture is same for the baseline model.}
For textual embeddings, we tested with and without attention layer (A0 and A1) and we varied the number of hidden state features in LSTM (HSx). For visual embeddings, we used ResNet-18~\cite{he2016deep} or MobileNetV2 (RES or MB) as the initial network, and either freezed the pretrained weights on ImageNet (FR), trained the network with randomly initialized weights (RI), or fine-tuned the pretrained weights (FT). For multimodal embeddings, we used five fusion methods explained in the previous section, which are concatenation, sum, weighted sum with scalar weight, weighted sum with weight vector and weighted sum with weight obtained from a network, enumerated in this order.
% and the hash value is used as the input of a linear layer with the same output width as our other embedding methods. Our goal is to obtain embedding vectors of similar size with other methods by just using the sequentiality information. As no tasks in the same vein of our semantic parsing goal has not been covered in the existing literature, we essentially create this baseline to compare our proposed embedding methods against a minimum level of performance, where the models only access the locations of pages inside a document and do not use the pages' content (i.e. visual and textual features).

\begin{table}[ht]
\centering
\caption{Results for textual, visual and multimodal fusion embeddings. Best result for each modality is given in bold. Model abbreviations is explained under the experimental results subsection.}
\scalebox{0.95}{
\begin{tabular}{l|ll|ll|ll}
\hline
\multicolumn{1}{c|}{\textbf{Embedding}} & \multicolumn{2}{c|}{\textbf{\begin{tabular}[c]{@{}c@{}}Page Stream\\ Segmentation\end{tabular}}} &  \multicolumn{2}{c|}{\textbf{\begin{tabular}[c]{@{}c@{}}Semantic\\ Parsing\end{tabular}}} & \multicolumn{2}{c}{\textbf{Classification}} \\ \hline
 & \multicolumn{1}{c}{ACC} & \multicolumn{1}{c|}{F1} & \multicolumn{1}{c}{UAS} & \multicolumn{1}{c|}{LAS} & \multicolumn{1}{c}{ACC} & \multicolumn{1}{c}{F1} \\\hline
%BASELINE & 0.566 & 0.704 & 0.485 & 0.302 & 0.274 & 0.071 \\ 
BASELINE & 0.566 & 0.704 & 0.485 & 0.302 & 0.274 & 0.071 \\ 
%baseline & \textbf{0.819} & \textbf{0.861} & \textbf{0.778} & \textbf{0.706} & \textbf{0.994} & \textbf{0.775} \\ 
\hline
 \textbf{TEXTUAL} &  & &  &  &  &  \\ 
A1-HS32 & 0.819 & 0.872 & 0.709 & 0.628 & 0.671 & 0.362 \\
A1-HS64 & \textbf{0.837} & \textbf{0.884} & 0.710 & 0.633 & \textbf{0.690} & 0.363 \\ 
A0-HS128 & 0.822 & 0.878 & 0.704 & 0.615 & 0.656 & 0.347 \\ 
A1-HS128 & 0.833 & 0.881 & \textbf{0.724} & \textbf{0.642} & 0.686 & \textbf{0.373} \\ 
\hline
\textbf{VISUAL} &  & &  &  &  &  \\ 
RES-FR & 0.863 & 0.900 & 0.690 & 0.611 & 0.641 & 0.332 \\ 
MB-FR & 0.872 & 0.906 & 0.720 & 0.652 & 0.665 & 0.331 \\ 
MB-RI & 0.882 & 0.908 & 0.751 & 0.684 & 0.650 & 0.329 \\ 
MB-FT & \textbf{0.898} & \textbf{0.925} & \textbf{0.766} & \textbf{0.699} & \textbf{0.683} & \textbf{0.363} \\ 
\hline
\textbf{FUSION} &  & &  &  &  &  \\ 
MM1 & 0.890 & 0.921 & 0.750 & 0.684 & \textbf{0.741} & 0.408 \\ 
MM2 & 0.894 & 0.922 & \textbf{0.776} & \textbf{0.712} & 0.740 & 0.413 \\ 
MM3 & \textbf{0.895} & \textbf{0.922} & 0.765 & 0.702 & 0.738 & 0.403 \\ 
MM4 & 0.882 & 0.912 & 0.766 & 0.695 & 0.739 & \textbf{0.426} \\ 
MM5 & 0.894 & 0.921 & 0.767 & 0.704 & 0.740 & 0.425 \\ 
\hline
\end{tabular}
}
\label{Table:Result_Embedding}
\end{table}

\begin{table}[ht]
\centering
\caption{F1 scores for each dependency label for semantic parsing.}
\scalebox{0.95}{
\begin{tabular}{l|lllll}
\hline
\multicolumn{1}{c|}{\textbf{Embedding}} & \multicolumn{1}{c}{\textbf{atch}} & \textbf{copy} & \multicolumn{1}{c}{\textbf{next}} & \textbf{back} & \textbf{root} \\ \hline
BASELINE & 0.125 & 0.135 & 0.511 & 0.090 & 0.723 \\ \hline
\textbf{TEXTUAL} &  &  &  &  &  \\
A1-HS32 & 0.665 & 0.783 & 0.775 & 0.775 & 0.844 \\
A1-HS64 & 0.666 & \textbf{0.790} & \textbf{0.775} & \textbf{0.791} & \textbf{0.879} \\ 
A0-HS128 & 0.648 & 0.786 & 0.771 & 0.742 & 0.844 \\ 
A1-HS128 & \textbf{0.677} & 0.789 & 0.769 & 0.779 & 0.876 \\ 
\hline
\textbf{VISUAL} &  &  &  &  &  \\
RES-FR & 0.633 & 0.740 & 0.784 & 0.869 & 0.816 \\ 
MB-FR & 0.679 & 0.789 & 0.812 & 0.905 & 0.849 \\ 
MB-RI & 0.702 & 0.811 & 0.820 & \textbf{0.909} & \textbf{0.868} \\ 
MB-FT & \textbf{0.713} & \textbf{0.813} & \textbf{0.842} & 0.897 & 0.845 \\ 
\hline
\textbf{FUSION} &  &  &  &  &  \\
MM1 & 0.698 & 0.810 & 0.830 & 0.890 & 0.847 \\ 
MM2 & 0.733 & 0.809 & 0.834 & 0.906 & 0.894 \\ 
MM3 & \textbf{0.734} & \textbf{0.825} & \textbf{0.838} & 0.907 & 0.930 \\ 
MM4 & 0.726 & 0.811 & 0.817 & 0.900 & 0.874 \\ 
MM5 & 0.700 & 0.817 & 0.832 & \textbf{0.911} & \textbf{0.933} \\   \hline
\end{tabular}
}
\label{Table:Result_Dependency}
\end{table}

From the experimental results, we can observe that the proposed approach significantly improves the baseline even with unimodal embeddings. Both textual and visual embeddings provide more than $100\%$ increase in LAS for semantic parsing, which is our main task. Furthermore, it can be seen that visual features are of utmost importance in both semantic parsing and page stream segmentation, as the best performing visual method outperformed the textual method by over 6 percentage points in both page stream segmentation accuracy and semantic parsing LAS. Results in classification task are much similar with only a difference of 1.3 percentage points in accuracy. This shows the importance of visual patterns in semantic parsing compared to regular classification tasks.

For textual embeddings, we see that increasing hidden size of the LSTM layers slightly improves semantic parsing performance, with no clear trend in other two tasks. For visual embeddings, MobileNet architecture outperforms ResNet when both networks are used with pretrained weights. We consider this behavior as a result of our rather small dataset, which leads to MobileNet, which has 3.4M parameters, to be trained more effectively compared to ResNet with 11.5M parameters. Performance of MobileNet increases moderately when we start with randomly initialized weights and train the model from scratch, but best performance is obtained with pretrained weights fine-tuned on our dataset. As the models were trained on ImageNet, fine-tuning adapts models to the document domain, confirming findings from the previous works~\cite{engin2019multimodal}. Based on the findings for each modality, we use A1-HS128 and MB-FT configurations for our multimodal embedding setup.

It can be observed that the main contribution of multimodal approach comes in classification task. Fusion embeddings bring similar improvements over textual embeddings, but results are very similar or slightly lower than best performing visual embeddings in page stream segmentation and semantic parsing, which was also observed in other works comparing visual and multimodal features~\cite{audebert_multimodal_2019}. However, we see that multimodality improves the classification accuracy by 5 percentage points over unimodal approaches. We explain this with the classification task containing classes that have distinct visual structure as well as classes that are made up of freeform documents. While unimodal embeddings can specialize in only one of these categories, using multimodal information enables us to both capture visual patterns and utilize details from the freeform text.  We also do not see any clear tendency towards a particular fusion method, but MM2, which is elementwise addition, outperforms the closest method by 1 percentage point in semantic parsing. Again, the lack of a significant improvement when using learnable weights can be due to the limitations on the amount of training data. 
  
Differences between textual and visual features are more apparent in Table~\ref{Table:Result_Dependency}, where F1 scores for each dependency label is presented. Here, we see that textual embeddings diverges negatively in predicting \texttt{back} and \texttt{next} labels by 11.8 and 6.7 percentage points from the visual embeddings, respectively. We attribute this difference to the semantic structure of the labels. Back pages are easily identifiable from the visuals, whereas small footers that do not take up much space on the page can be detected by OCR and confuse the text page. Similarly, next pages often share similar visual patterns with the head page, which cannot be detected by OCR systems. Moreover, it can be seen that multimodal embeddings improve upon unimodal approaches in 4 of 5 labels, only falling short of visual embeddings in \texttt{next} label by $0.5$ percentage points. This highlights the importance of different modalities in parsing semantic relations.

\section{Conclusion}
In this work, we proposed a multi-task approach for document processing that parses semantic relations between pages in addition to page stream segmentation and classification using multimodal feature representations. Our work is the first study to focus on parsing semantic interpage relations. Our experimental results show that our method achieves satisfactory performance in our novel task, outperforming unimodal approaches. Based on our experiments on international trade documents, we show that this framework can replace consistency checks traditionally done by human operators and increase efficiency of digitization workflows. 

%\clearpage


%\section*{Acknowledgment}




\bibliographystyle{IEEEtran}

\bibliography{IEEEabrv,root}
%\begin{thebibliography}{1}


%\end{thebibliography}

\end{document}

\title{Supplementary Material for\\Semantic Parsing of Interpage Relations}

%\author{\IEEEauthorblockN{Mehmet Arif Demirta\c{s}}
%\IEEEauthorblockA{Department of Computer Engineering\\
%Istanbul Technical University\\
%Istanbul, Turkey\\
%Email: demirtasm18@itu.edu.tr}
%\and
%\IEEEauthorblockN{Berke Oral}
%\IEEEauthorblockA{Yap{\i} Kredi Teknoloji\\
%Istanbul, Turkey\\
%Email: berke.oral@ykteknoloji.com.tr}
%\and
%\IEEEauthorblockN{Mehmet Yasin Akpınar}
%\IEEEauthorblockA{Yap{\i} Kredi Teknoloji\\
%Istanbul, Turkey\\
%Email: mehmetyasin.akpinar@ykteknoloji.com.tr}
%\and
%\IEEEauthorblockN{Onur Deniz}
%\IEEEauthorblockA{Yap{\i} Kredi Teknoloji\\
%Istanbul, Turkey\\
%Email: onur.deniz@ykteknoloji.com.tr}}


%\author{\IEEEauthorblockN{Mehmet Arif Demirta\c{s}\IEEEauthorrefmark{1}\IEEEauthorrefmark{2},
%Berke Oral\IEEEauthorrefmark{2},
%Mehmet Yasin Akpınar\IEEEauthorrefmark{2} and
%Onur Deniz\IEEEauthorrefmark{2}}
%\IEEEauthorblockA{\IEEEauthorrefmark{1}Department of Computer Engineering, Istanbul Technical University, Istanbul, Turkey}
%\IEEEauthorblockA{\IEEEauthorrefmark{2}Yap{\i} Kredi Teknoloji, Istanbul, Turkey}
%\IEEEauthorblockA{Email: demirtas18@itu.edu.tr, \{berke.oral, mehmetyasin.akpinar, onur.deniz\}@ykteknoloji.com.tr}}

\DeclareRobustCommand*{\IEEEauthorrefmark}[1]{%
  \raisebox{0pt}[0pt][0pt]{\textsuperscript{\footnotesize\ensuremath{#1}}}}
\author{\IEEEauthorblockN{Mehmet Arif Demirta\c{s}\IEEEauthorrefmark{1,}\IEEEauthorrefmark{2},
Berke Oral\IEEEauthorrefmark{2},
Mehmet Yasin Akpınar\IEEEauthorrefmark{2} and
Onur Deniz\IEEEauthorrefmark{2}}
\IEEEauthorblockA{\IEEEauthorrefmark{1}Department of Computer Engineering, Istanbul Technical University, Istanbul, Turkey}
\IEEEauthorblockA{\IEEEauthorrefmark{2}Yap{\i} Kredi Teknoloji, Istanbul, Turkey}
\IEEEauthorblockA{Email: \{demirtas18, oralbe\}@itu.edu.tr, \{mehmetyasin.akpinar, onur.deniz\}@ykteknoloji.com.tr}}


\maketitle


\vspace{1em}

% \fi
\IEEEpeerreviewmaketitle

\section{Appendix A}
% This demo file is intended to serve as a ``starter file''
% for IEEE conference papers produced under \LaTeX\ using
% IEEEtran.cls version 1.8b and later.
% I wish you the best of success.



% The verification step consists of checking if the listed subdocuments are sent in the correct counts. 

\begin{figure}[h]
    \centering
    \includegraphics[width=0.50\textwidth, trim={0 3.5cm 7cm 2cm}, clip]{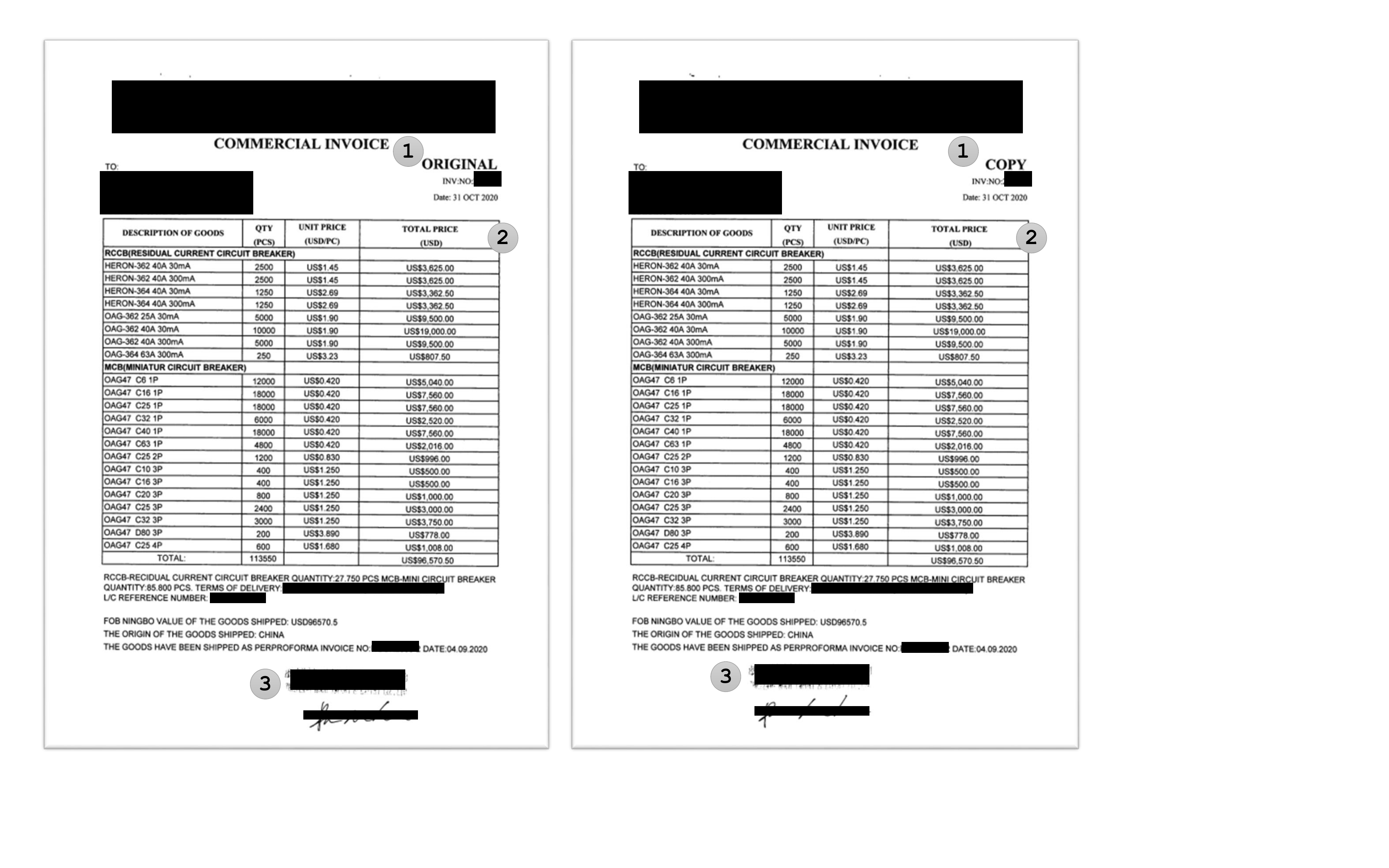}
    \caption{An example of similar visual layout between two copy pages. Both pages share the same header except for the tag ORIGINAL/COPY (1), both include a complex table (2) and the stamp and signature of the authorized figure in slightly different positions (3).}
    \label{fig:VisualLayout}
\end{figure}

As mentioned in the main text, visual and textual features obtained from pages may provide unique contributions to the model, as both modalities include valuable information from the page. Figure \ref{fig:VisualLayout} shows a case where two pages share similar visual layouts, but textual content differs.

The difference in textual content, stemming from noise in OCR processes, can be mainly categorized under two factors. The first one is the erroneous characters, which will lead to a completely different word vector. The second one is caused by the nature of the used OCR product. Since the recognition is done in a block by block basis, i.e. the page to be recognized is segmented first and each block is recognized afterwards, the blocks are often segmented in a different order and/or content. For example, if we look at the snippets from the OCR results of the pages presented in Figure \ref{fig:VisualLayout}, the outcome is as follows:
\begin{itemize}
    \item Original Page:\\
    RCCB(RESIDUAL CURRENT CIRCUIT BREAKER) HERON-362 40A 30mA 2500 US\$1.45 US\$3,625.00 HERON-362 40A 300mA 2500 US\$1.45 US\$3,625.00 HERON-364 40A 30mA 1250 US\$2.69 US\$3,362.50 HERON-364 40A 300mA 1250 US\$2.69 US\$3,362.50 OAG-362 25A 30mA 5000 US\$1.90 US\$9,500.00\\
    OAG-362 40A 30mA 10000 US\$1.90 US\$19,000.00\\
    OAG-362 40A 300mA 5000 US\$1.90 US\$9,500.00\\
    OAG-364 63A 300mA 250 US\$3.23 US\$807.50
    \item Copy Page:\\
    RCCB(RESIDUAL CURRENT CIRCUIT BREAKER) HERON-362 40A 30mA 2500 US\$1.45\\
    HERON-362 40A 300mA 2500 US\$1.45\\
    HERON-364 40A 30mA 1250 US\$2.69\\
    HERON-364 40A 300mA 1250 US\$2.69\\
    OAG-362 25A 30mA 5000 US\$1.90\\
    OAG-362 40A 30mA 10000 US\$1.90\\
    OAG-362 40A 300mA 5000 US\$1.90\\
    OAG-364 63A 300mA 250 US\$3.23\\ US\$3,625.00 US\$3,625.00 US\$3,362.50 US\$3,362.50 US\$9,500.00 US\$19,000.00 US\$9,500.00 US\$807.50
\end{itemize}
This situation results in a different ordering of the overall text, which would result in a low similarity score if we were doing character comparison, even though the pages looks close to identical to the human eye.

\section{Appendix B}
For extracting textual content, the OCR software used is ABBYY FineReader Engine, Version 11 Release 8. The parameters used are listed below:
\begin{itemize}
    \item CorrectInvertedImage: true
    \item FastMode: true
    \item LowResolutionMode: true
    \item DetectTables: true
    \item EnableTextExtractionMode: true
    \item ProhibitModelAnalysis: true
    \item DetectPictures: false
    \item DetectTextOnPictures: true
    \item TextTypes: TT\_Normal $\vert$ TT\_Typewriter $\vert$ TT\_OCR\_B
    \item GeometryCorrectionMode: GCM\_Auto
    \item CorrectSkew: True
    \item CorrectSkewMode: CSM\_CorrectSkewByHorizontalLines $\vert$ CSM\_CorrectSkewByHorizontalText
\end{itemize}
The rest of the parameters are left in default values.